\pdfoutput=1
\documentclass[11pt]{article}
\usepackage{acl}
\usepackage{times}
\usepackage{latexsym}
\usepackage[T1]{fontenc}
\usepackage[utf8]{inputenc}
\usepackage{microtype}
\usepackage[ruled,vlined]{algorithm2e}
\usepackage{amsfonts}
\usepackage{multirow}
\usepackage{inconsolata}

\usepackage{booktabs}
\usepackage{tabularx}
\usepackage{xspace}
\newcommand{\emoclass}{\texttt{$\langle$em$\rangle$}\xspace}
\newcommand{\F}{$\textrm{F}_1$\xspace}

\title{Emotion-Conditioned Text Generation through\\ Automatic Prompt Optimization}

\author{Yarik Menchaca Resendiz \and Roman Klinger\\
  Institut f\"ur Maschinelle Sprachverarbeitung, University of Stuttgart\\
  \texttt{\{yarik.menchaca-resendiz,roman.klinger\}@ims.uni-stuttgart.de}
}

\begin{document}
\maketitle
\begin{abstract}
  Conditional natural language generation methods often require either
  expensive fine-tuning or training a large language model from
  scratch. Both are unlikely to lead to good results without a
  substantial amount of data and computational resources. Prompt
  learning without changing the parameters of a large language model
  presents a promising alternative.  It is a cost-effective approach,
  while still achieving competitive results. While this procedure is
  now established for zero- and few-shot text classification and
  structured prediction, it has received limited attention in
  conditional text generation. We present the first automatic prompt
  optimization approach for emotion-conditioned text generation with
  instruction-fine-tuned models. Our method uses an iterative
  optimization procedure that changes the prompt by adding, removing,
  or replacing tokens. As objective function, we only require a text
  classifier that measures the realization of the conditional variable
  in the generated text. We evaluate the method on emotion-conditioned
  text generation with a focus on event reports and compare it to
  manually designed prompts that also act as the seed for the
  optimization procedure. The optimized prompts achieve 0.75
  macro-average \F to fulfill the emotion condition in contrast to
  manually designed seed prompts with only 0.22 macro-average \F.
 \end{abstract}

\section{Introduction}
\label{sec:introduction}
Emotions are fundamental in communication, where they play an
important role in transferring meaning and intent
\cite{ekman1992argument}. Emotion-conditioned natural language
generation models aim at improving human--computer interaction, by
generating text that is not limited to conveying propositional
information.  However, state-of-the-art conditional generation models
require a large amount of data and computational power to achieve
models that allow for a
fine-grained control over the generated texts
\cite{pascual-etal-2021-plug-play, ghosh-etal-2017-affect,
  song-etal-2019-generating, ECM, MenchacaResendiz2023}.

In areas like text classification or structured prediction, prompt
optimization has established itself as a zero- or few-shot learning
paradigm \citep{ding2021openprompt, zhang-etal-2022-prompt-based,
  wang-etal-2022-towards-unified}, also in emotion analysis
\citep{Plazadelarco2022,zheng-etal-2022-ueca,yin-etal-2019-benchmarking}.
Here, only parameters that are concatenated to the input are optimized
and the large language model's parameters are frozen. Such models,
therefore, exploit encoded knowledge in models such as Flan
\citep{tay2023ul2}, GPT-3 \citep{brown2020language} and Alpaca
\cite{alpaca} more explicitly than fine-tuning them for the task at
hand. The optimization method learns ``how to use'' a model, not ``how
to change'' it.

\begin{table}
  \centering\footnotesize
  \setlength{\tabcolsep}{4pt}
  \begin{tabularx}{\linewidth}{lXp{31mm}}
    \toprule
    I. & Input prompt & Generated text\\
    \cmidrule(){1-1} \cmidrule(lr){2-2} \cmidrule(l){3-3}
    0 & \texttt{Text with disgust} &  \texttt{Disgust is a character in Inside Out}\\
    \cmidrule(){1-1} \cmidrule(lr){2-2} \cmidrule(l){3-3}
    1 & \texttt{Text expressing disgust} & \texttt{Disgusting}  \\
    \cmidrule(){1-1} \cmidrule(lr){2-2} \cmidrule(l){3-3}
    2 & \texttt{Write a text to express disgust} & \texttt{A look of disgust came over his face.} \\
    \bottomrule
  \end{tabularx}
  \caption{Hypothetical example for a prompt optimization process. The
    seed prompt is given in Iteration (I.) 0 and misinterpreted to mention
    the character ``Disgust''. This issue is fixed through iterative
    optimization.\\[-2pt]}
  \label{tab:methodoly_goal}
\end{table}

In recent instruction-based models, the prompt is an instruction to
elicit a desired response. The instruction serves as a starting point
for generating text that aligns with the intended task.  Prompting in
text classification \cite{hu-etal-2022-knowledgeable,
  gu-etal-2022-ppt} usually includes the instruction (e.g., ``classify
the text\ldots'') and the label representation (e.g., ``positive'',
``negative''). Summarization has been represented as an instruction by
appending ``TL;DR'' or ``summarize'' \cite{radford2019language,
  narayan-etal-2021-planning}. For text generation that translates
tables to text, \citet{li-liang-2021-prefix} proposed to tune a prefix
prompt to accomplish the task. In machine translation, prompts
typically mention the source and target language, such as ``translate
English to German'' \cite{raffel2019exploring}.

The task of prompt optimization can be formulated in various
directions. The goal is to find the optimal sequence of tokens to
represent the prompt for a specific model (e.g., Flan) and task (e.g.,
summarization), while keeping the model weights unchanged. AutoPrompt
\cite{autoprompt:emnlp20} defines the prompt optimization as
``fill-in-the-blanks'' based on a gradient-guided search. OpenPrompt
\cite{ding2021openprompt} provides a toolkit for training prompts
using a template dataset, along with corresponding verbalizers for
different classes. \newcite{deng-etal-2022-rlprompt} use reinforcement
learning to infer a successful prompt variation strategy. A different
approach for optimization is fine-tuning the model to improve its
performance with a specific prompt, while keeping the prompt unchanged
\cite{jian-etal-2022-contrastive, gu-etal-2022-ppt}.

In contrast to most previous work, we use models that have been
fine-tuned to solve instruction-based tasks; in our case to generate
emotion-conditioned texts. This comes with distinct challenges because
the loss function cannot be determined by a single expected label
(e.g., positive or negative). In our work, we use a classifier that
measures the fulfillment of the condition as a source to calculate the
value of an objective function. The optimization procedure that we
propose is an evolutionary optimization method \citep{Simon2013}. Next
to the objective function, an important component are actions that
allow changes to a prompt to explore the search space.

\section{Methods}
\label{sec:methods}
We propose a method (summarized in pseudocode in
Algorithm~\ref{algoritm:pseudocode}) for text generation conditioned
on emotions using prompt optimization. It involves an iterative
optimization procedure with three modules, namely \textit{prompt
  modification}, \textit{text generation}, and \textit{prompt
  evaluation}. We describe the modules in Section \ref{subsec:modules}
and the iterative optimization in Section~\ref{sec:pseudocode}.

\subsection{Modules}
\label{subsec:modules}

\paragraph{Prompt modification.}
In each optimization iteration, we apply the three operations,
one at a time, to all the tokens in the prompt. Therefore, based on
one ``parent'' prompt, we create $\lambda>1$ ``children''.

\textit{Addition} adds the most probable token at any position within
the prompt, including both the beginning and end of the prompt. We use
the pre-trained RoBERTa model \citep{liu2019roberta} to retrieve
probable tokens for each of these positions.
\textit{Removal} deletes a token from the prompt.
The \textit{Replacement} operation exchanges a token by
the most probable token, again as predicted by RoBERTa.

The \textit{Addition}
and \textit{Replacement} operations use the \texttt{<mask>} special token to
predict the word. We exemplify these operations in Table
\ref{tab:operation_examples}.

\begin{table}
  \centering\small
  \setlength{\tabcolsep}{5pt}
  \begin{tabular}{lll}
    \toprule
    Original Prompt & Oper. & Modified Prompt \\
    \cmidrule(r){1-1} \cmidrule(lr){2-2} \cmidrule(l){3-3}
    Text that expresses & Add. & Text \underline{string} that expresses \\
    Text \textbf{that} expresses & Repl.& Text \underline{a} expresses \\
    Text \textbf{that} expresses & Rem. & Text expresses \\
    \bottomrule
  \end{tabular}
  \caption{The prompt operations (Oper.) are performed on the same prompt. The Addition (Add.) adds RoBERTa's special mask token (\texttt{<mask>}) between \textit{Text} and \textit{that}. The Replacement (Repl.) masks the target word (that). The unmasked/predicted tokens by RoBERTa are \underline{underlined}, and the replaced or removed tokens from the original are in \textbf{bold}. Removal (Rem.) deletes one token from the prompt.}
  \label{tab:operation_examples}
\end{table}

\paragraph{Text generation.}
We then use each of the $\lambda$ prompt variations to create text
using a large pre-trained language model (e.g., Flan). To do so, we
instantiate it with the emotion category. We refer to this
instantiation as the \textit{Conditional-Prompts}. Each of them
consists of the modified prompt and the specified emotion (e.g.,
``Text that expresses \emoclass''). Here, \emoclass is replaced by
each of the emotion categories under consideration.

\paragraph{Evaluation.}
Each prompt is then evaluated through the texts that are generated
with its instantiated \textit{Conditional-Prompts}. In the evaluation,
we do not further consider texts that are a paraphrase of the
Conditional-Prompt. We calculate the BLEU score
\citep{papineni-etal-2002-bleu} and filter all texts with a score
greater than 0.2. For example, a language model could generate ``The
text expresses joy.'' for a Conditional-Prompt ``Text that expresses
joy''.

The actual evaluation is performed by comparing the emotion condition
to the judgment of an emotion classifier, applied to the generated
texts. We use the \F measure both as an objective function during
optimization and for final evaluation. Note that these two scores are
based on two separate classifiers, trained on independent data.

\subsection{Iterative Optimization}
\label{sec:pseudocode}

Algorithm \ref{algoritm:pseudocode} shows the iterative prompt
optimization for a given seed prompt $P$ (e.g., ``Text that
expresses''). The optimization is based on a $(\mu, \lambda)$
evolutionary algorithm \cite{eiben2015introduction}, more concretely
$(1, \lambda)$, because we keep only the one best-performing prompt
for the next optimization iteration. In contrast to a $(\mu+\lambda)$,
the respective parent is not further considered in the next
iteration. This makes the algorithm less likely to get stuck in a
local optimum.

Initially, $P_{\textit{opt}}$ (the optimized prompt) is initialized
with the seed prompt $P$. Next, each token in $P_\textit{opt}$ is
modified using the Addition, Replacement, and Removal. Each operation
is performed one at a time, and the results are stored in
$\mathbf{P}_\textit{mod}$ (Section \ref{subsec:modules}).  The
\textit{Generate} method produces a text for each
\textit{Conditional-Prompt}-combination of the input prompt and the
emotion class (e.g., ``Text that expresses joy'', ``Text that
expresses anger''; Section \ref{subsec:modules}). We compare the
generated text from $P_\textit{opt}$ (namely $T_\textit{opt}$) against
the generated text from each modified prompt
($\mathbf{P}_\textit{mod}$), denoted as $\mathbf{T}_\textit{mod}$. If
the \F of $\mathbf{T}_\textit{mod}$ is higher than that of
$T_\textit{opt}$, the prompt $\textit{prompt}_\textit{mod}$ is
assigned as the new optimized prompt ($P_\textit{opt}$) and added to
the best-performing candidates ($\mathbf{P}_{\textit{cands}}$).
Finally, this process is repeated for a total of $N$ times and
$P_\textit{opt}$ is updated with the best-performing prompt from
$\mathbf{P}_{\textit{cands}}$.

\begin{algorithm}[t]
  \small
  \SetKwInOut{Input}{Input} \SetKwInOut{Output}{Output}
  \Input{Seed Prompt $P$, \linebreak Maximum Iterations $N$}
  \Output{Optimized Prompt $P_\textit{opt}$}
  \SetKwFunction{Add}{Add}
  
  $P_\textit{opt} \gets P$\;
  $i \gets 0$\;
  $\mathbf{P}_\textit{cands} \gets \{\}$\;
  
  \While{$i < N$}{
    $\mathbf{P}_\textit{mod} \gets \{\}$\; 
    
    \For{$\textit{token} \in P_\textit{opt}$}{
      $\mathbf{P}_\textit{mod}\,+\kern-3pt=\textit{Add}(P_\textit{opt}, \textit{token})$\;
      $\mathbf{P}_\textit{mod}\,+\kern-3pt=\textit{Replace}(P_\textit{opt}, \textit{token})$\;
      $\mathbf{P}_\textit{mod}\,+\kern-3pt=\textit{Remove}(P_\textit{opt}, \textit{token})$\;}
    
    $\mathbf{T}_\textit{opt} \gets \{\}$;
    
    \For{$\textit{prompt}_\textit{mod} \in \mathbf{P}_\textit{mod}$}{
      $\mathbf{T}_\textit{mod} \gets \textit{Generate}(\textit{prompt}_\textit{mod}) $\;
      
      \If{$\textit{Eval}(\mathbf{T}_\textit{mod}) > \textit{Eval}(\mathbf{T}_\textit{opt})$}{
        $P_\textit{opt} \gets \textit{prompt}_\textit{mod}$\;
        $\mathbf{T}_\textit{opt} \gets \mathbf{T}_\textit{mod}$\;
      }  	
    }
    $\mathbf{P}_\textit{cands} +\kern-3pt= \textit{P}_\textit{opt}$

    $i \gets i + 1$\;
  }
  $P_\textit{opt} \gets \textrm{select-one-best}(\mathbf{P}_\textit{cands})$\;
  
  \Return $P_\textit{opt}$\;
  \caption{Automatic Prompt Optimization. \textit{Eval} involves an emotion classifier and the BLEU score.}
  \label{algoritm:pseudocode}
\end{algorithm}

\section{Experiments}
\label{experiments}

Section \ref{subsec:experimental_settings} explains the experimental
settings used to optimize an initial prompt that we assume to be
provided by a user. Section \ref{subsec:resutls}
validates the proposed method by showing that emotion-conditioned text
generation improves when using the optimized prompt compared to the
seed prompt.

\subsection{Experimental Settings}
\label{subsec:experimental_settings}
To validate the feasibility of our method for emotion-conditioned text
generation, and its cost-effectiveness in terms of data and
computational resources, we utilized available pre-trained models and
datasets. Specifically, we used Flan \cite{tay2023ul2}, an open-source
model trained on instruction-based datasets, as a generative model. We
trained two classifiers using (1) the ISEAR dataset
\cite{scherer1994evidence} for prompt optimization in each iteration,
and (2) the crowd-enVent dataset \cite{Troiano2023} for final
evaluation, utilizing the same subset of emotions as the ISEAR
dataset.\footnote{The emotion labels are: Anger, Disgust, Fear, Guilt,
  Joy, Sadness, and Shame.} Both classifiers are built on top of 
RoBERTa using default parameters for 10
epochs.\footnote{The crowd-enVent and ISEAR-based classifiers have
  macro-\F of .78 and .77, respectively.}

These data sets are independent of each other, and therefore the
objective signal is independent of the final evaluation. Both sets,
however, are comparable: they contain texts in which people were asked
to report on an emotion-triggering event, given a predefined
emotion. In the original ISEAR corpus, these texts were acquired in an
in-lab setting in the 1990s, while the crowd-enVENT corpus has
recently been collected in 2022 in a crowd-sourcing setup. An example
from the ISEAR corpus is ``When I was involved in a traffic
accident.'' -- an example from crowd-enVENT is ``When my son was
poorly with covid''.

\paragraph{Prompt Modification.} We selected a straightforward seed
prompt---``Write a text that expresses \emoclass''---for ten
iterations and all operations.

\paragraph{Text Generation.} For each \textit{Conditional-Prompt}, we generate the three most
probable sentences using a beam search with a beam size of 30, a next-token temperature
of 0.7, and a top-p (nucleus) sample of 0.7. We ensure that our output excludes
sentences with repeated instances of the same bigram.

\paragraph{Prompt Evaluation.} We filter out all prompts where the average
BLEU score is higher than 0.2 across all the conditional
prompts. Next, we select the prompt with the best \F score using the ISEAR
classifier.

\subsection{State-of-the-art Baseline}
\label{subsec:sota}
We compare our method against the plug-and-play method proposed by
\citet{pascual-etal-2021-plug-play}---a state-of-the-art model for
affective text generation. To do so, we train the emotion
discriminators that are part of that method on top of GPT-2 with the
ISEAR dataset. The comparison is not straightforward since this method
uses the prompt as a starting point to generate the sentence, whereas
our approach treats the prompt as an instruction. Therefore, we select
the most frequent n-grams from the ISEAR dataset as prompts: ``When I
was'', ``When a'', and ``When someone''. For each
prompt--discriminator combination, we generate the 5 most probable
sentences.

\begin{table}
  \centering\small
  
  \setlength{\tabcolsep}{2pt}
  \begin{tabularx}{\linewidth}{lcXr}
    \toprule
    I. & Ope. & Optimized Prompt ($P_{opt}$) & \F \\
    
    \cmidrule(r){1-1} \cmidrule(rl){2-2} \cmidrule(lr){3-3} \cmidrule(l){4-4}
    
    0 & --- & Write a text that expresses \emoclass & .28 \\
    1 & Repl. & Write a text to expresses \emoclass & .80  \\
    2 & Add. & Write in a text to expresses \emoclass & .91\\
    3 & Add. & Write in a text string to expresses \emoclass & .88 \\
    4 & Add. & Write in a long text string to expresses \emoclass & \textbf{.94} \\
    5 & Rem. & Write in long text string to expresses \emoclass & \textbf{.94} \\
    6 & Repl. & Write in long text strings to expresses \emoclass & .91 \\
    
    \bottomrule
  \end{tabularx}
  \caption{Prompt optimization at different iterations (I.), with
    Iteration 0 representing the seed prompt. The \emoclass token
    represents any of the seven emotions in the ISEAR dataset. The
    macro \F score is calculated using the ISEAR classifier, across
    all the emotions.}
  \label{tab:prompt_optimization}
\end{table}

\subsection{Results}
\label{subsec:resutls}
\begin{table*}
  \centering\small
  \setlength{\tabcolsep}{5pt}
  \begin{tabular}{lllr}
    \toprule
    Iter. & Prompt & Generated text & \F \\
    
    \cmidrule(l){1-1} \cmidrule(l){2-2} \cmidrule(l){3-3} \cmidrule(l){4-4}
    
    0 & Write a text that expresses disgust & i don't know what to do about it. & .28\\
    1 & Write a text to expresses disgust & i'm disgusted. & .80 \\
    1 & a text that expresses fear &  a text that expresses fear & .90 \\
    2 & Write in a text to expresses shame & i don't know how to express my shame. & .91 \\
    2 & Write a text that to expresses shame & i'm sorry to hear that. & .58 \\
    5 & Write in a long enough string to expresses joy & a long enough string to express joy. & 1.0 \\
    5 & Write a long text string to expresses joy  & i love you so much & \textbf{.94} \\
    
    \bottomrule
  \end{tabular}
  \caption{Generated text examples from the same seed prompt (1st row)
    at different optimization steps. The macro \F score is computed
    for the prompt across all the emotions using the ISEAR
    classifier.}
  \label{tab:prompt_examples}
\end{table*}

We begin the discussion of the results with
Table~\ref{tab:prompt_optimization}, which shows the prompt
optimization and performance across iterations. It reveals two notable
findings:
First, already the first iteration, compared to the seed prompt in
Iteration 0, shows an increase by 52\,pp in \F.  This is a change only
by replacing ``that'' with ``to''.  Given our selection criteria which
does not include the parent prompt in the selection, there can be a
decrease in performance, which can be observed in Iteration 2.
Second, all prompts in Table \ref{tab:prompt_optimization}---the best-performing
prompts at each iteration---are human-readable. This is in
contrast to prompt optimization in other NLP tasks, where the
resulting prompts often become less human-readable. For example, in
the fact retrieval task ``[X] limestone depositedati boroughDepending
[Y]'' performs better than ``[X] is the capital of [Y]''
\cite{ding2021openprompt}.

Table \ref{tab:prompt_examples} showcases examples of generated texts
from various prompt candidates. The prompt candidates at the same
iteration are a few examples of the resulting prompt modifications as
described in Section \ref{sec:methods}.  The provided \F scores refer
to the performance of the prompt across the 7 emotions, not the
performance of the specific examples shown. Comparing the generated
text from the seed prompt (Row 1) and the first optimization (Row 2),
we observe a better fulfillment of the emotion \textit{disgust} for
the optimized prompt---the uncertainty expressed in Row 1 indicates
\textit{fear}. Prompt modifications at the same iteration have
different performances.  For example, in Iteration 2 (Rows 4/5), there
is a difference of 33\,pp in \F.  It is important to note that the
best \F score does not always indicate an improvement in fulfilling
the condition of the generated text. Sometimes, the best-scoring text
can be a paraphrase of the prompt, which may be falsely classified as
correct due to the presence of the emotion class name (e.g., Row
6/Iteration 5, Row 3/Iteration 2).

Finally, Table \ref{tab:vs_pp} shows an independent evaluation of the
method along with the results achieved with the method by
\citet{pascual-etal-2021-plug-play}. We report \F scores for the
ISEAR-based classifier used during the optimization process and the
independent crowd-enVENT-based classifier. The latter numbers
therefore constitute an independent evaluation result. We observe that
the numbers of both classifiers are comparable to each other. The
comparison to the baseline shows that our seed prompt performs on par
with Pascual's method (.18, .12, and .17 vs.\ .22, respectively). Our
optimized prompt, however, shows a higher performance (.75 \F).

\begin{table}
  \centering\small
  \setlength{\tabcolsep}{2pt}
  \begin{tabularx}{\linewidth}{p{10mm}Xrr}
    \toprule
    &&& crowd-\\
    Method & Prompt & ISEAR & enVent\\
    \cmidrule(l){1-1} \cmidrule(l){2-2} \cmidrule(l){3-3} \cmidrule(l){4-4}
    
    \multirow{3}{*}{\parbox{15mm}{Pascual \shortcite{pascual-etal-2021-plug-play}}} & When I was &  .18 & .18\\
           & When a \ & .43 & .12  \\
           & When someone & .21 & .17\\
    \cmidrule(l){1-1} \cmidrule(l){2-2} \cmidrule(l){3-3} \cmidrule(l){4-4}
    
    \multirow{2}{*}{$P_\textit{opt}$} &  Write a text that expresses \emoclass &  .28 & .22\\
           &  Write in long text string to expresses \emoclass & .94  & \textbf{.75}\\
    
    \bottomrule
  \end{tabularx}
  \caption{Comparison between our method ($P_\textit{opt}$) and Pascual \shortcite{pascual-etal-2021-plug-play}
    Rows 1--3 are the most frequent n-grams for the ISEAR dataset. The 4th row corresponds
    to the seed prompt, and the 5th row represents the optimized prompt. The macro-average
    \F-score for both ISEAR and crowd-enVent datasets is computed across all emotions.}
  \label{tab:vs_pp}
\end{table}

\section{Conclusion and Future Work}

In this study, we introduced the first automatic prompt optimization
method for text generation conditioned on emotions. Our approach
involved three token operations: addition, replacement, and
removal. We utilized a BLEU score and an automatic classifier to
filter and rank the modified prompts. We demonstrated that the
optimized prompts led to a higher fulfillment of the intended emotions
compared to the seed prompt, with a 53\,pp improvement in the \F
score. It is a cost-effective method in terms of both data and
resource requirements, while still achieving good results.

This leads to important future work. While our approach improves
emotion-conditioned text generation, there are several areas that need
to be explored further. First, we need to explore different search
techniques for prompt optimization (e.g., Beam search).  Second, it is
essential to compare the performance of the optimized prompts across
different domains to assess the generalizability of our method. Our
evaluation is arguably comparably narrow, with only one seed prompt
and one domain in which emotions are expressed.  Finally, it is
crucial to analyze our approach by comparing it against a fine-tuned
or trained model from scratch to evaluate its effectiveness and
efficiency.

Another interesting direction of research would be to study in more
detail how the expected domain of the generated texts (here: emotion
self-reports) might be in conflict with the emotion condition and how
that can be encoded in either the optimization process, the seed
prompt selection or the objective functions, or in combinations of
these parameters.

\section{Ethical Considerations \& Limitations}
The proposed method aims at optimizing prompts for conditional text
generation, particularly when conditioned on emotions. The generated
affective texts do not only serve as a source to study the
capabilities of large language models from a computational
perspective. We believe that they can also be of value to better
understand the representation of psychological concepts in
automatically generated text.  However, there are some risks
associated with the method if not used with care, primarily inherited
from the underlying language model. Optimized prompts could
potentially result in generating text that reinforces stereotypes or
marginalize certain groups. When dealing with the expression of
emotions, it is essential to exercise caution when employing these
models due to their potential impact on individuals.

A limitation in our evaluation and method is that we rely heavily on
the seed prompts. This can lead to fast convergence---if the seed
prompt is adequate for the task, the optimization process is more
likely to be successful. The optimization is based on a
$(\mu, \lambda)$ approach, which can be seen as a brute-force
search. However, alternative search algorithms may provide a more
efficient optimization of the prompt in terms of iterations.

Overall, the method has proven to be useful for text generation
conditioned on emotions. We invite people to keep the above limitations
in mind when considering the capabilities and applications of our method.

 \section*{Acknowledgements}
 This work has been supported by a CONACYT scholarship
 (2020-000009-01EXTF-00195) and by the German
 Research Council (DFG), project
 ``Computational Event Analysis based on Appraisal Theories for Emotion
 Analysis'' (CEAT, project number KL 2869/1-2).

\bibliography{biblio.bib}

\end{document}